\documentclass[journal]{./sty/IEEEtran}
\usepackage{graphicx}
\usepackage{cite}
\usepackage{amsmath}
\usepackage{url}
\usepackage{amssymb}
\usepackage{tabularx}
\usepackage{multirow}
\usepackage{threeparttable}
\usepackage{caption}
\usepackage{booktabs}
\usepackage{ulem}
\usepackage{bbding}
\usepackage[ruled,linesnumbered]{algorithm2e}
\usepackage{color}
\newcolumntype{L}[1]{>{\raggedright\arraybackslash}p{#1}}
\newcolumntype{C}[1]{>{\centering\arraybackslash}p{#1}}
\newcolumntype{R}[1]{>{\raggedleft\arraybackslash}p{#1}}

\newcommand{\etc}{\textit{etc}}

\newcommand{\settablefont}{\fontsize{7.5}{8.6}\selectfont}

\begin{document}

\title{UP-CrackNet: Unsupervised Pixel-Wise Road Crack Detection via Adversarial Image Restoration}

\author{Nachuan Ma,~\IEEEmembership{Graduate Student Member,~IEEE},
	Rui Fan,~\IEEEmembership{Senior Member,~IEEE}, and 
	Lihua Xie,~\IEEEmembership{Fellow,~IEEE }

\thanks{Manuscript received June 26, 2023; revised November 04, 2023, January 09, 2024, and April 18, 2024; accepted April 28, 2024. This research was supported by the Fundamental Research Funds for the Central Universities and Xiaomi Young Talents Program. \emph{(Corresponding author: Rui Fan)}}
\thanks{N. Ma and R. Fan are with the College of Electronics \& Information Engineering, Shanghai Research Institute for Intelligent Autonomous Systems, the State Key Laboratory of Intelligent Autonomous Systems, and Frontiers Science Center for Intelligent Autonomous Systems, Tongji University, Shanghai 201804, China. (e-mails: manachuan@163.com, rui.fan@ieee.org) }
\thanks{L. Xie is with the School of Electrical and Electronic Engineering, Nanyang Technological University, 50 Nanyang Avenue, Singapore 639798 (e-mail: elhxie@ntu.edu.sg).}
\thanks{
Color versions of one or more figures in this article are available at
https://doi.org/10.1109/TITS.2024.3398037.
}
\thanks{
Digital Object Identifier 10.1109/TITS.2024.3398037
}
}

\markboth{IEEE TRANSACTIONS ON INTELLIGENT TRANSPORTATION SYSTEMS}{}

\maketitle

\begin{abstract}
Over the past decade, automated methods have been developed to detect cracks more efficiently, accurately, and objectively, with the ultimate goal of replacing conventional manual visual inspection techniques. Among these methods, semantic segmentation algorithms have demonstrated promising results in pixel-wise crack detection tasks. However, training such networks requires a large amount of human-annotated datasets with pixel-level annotations, which is a highly labor-intensive and time-consuming process. Moreover, supervised learning-based methods often struggle with poor generalizability in unseen datasets. Therefore, we propose an unsupervised pixel-wise road crack detection network, known as UP-CrackNet. Our approach first generates multi-scale square masks and randomly selects them to corrupt undamaged road images by removing certain regions. Subsequently, a generative adversarial network is trained to restore the corrupted regions by leveraging the semantic context learned from surrounding uncorrupted regions. During the testing phase, an error map is generated by calculating the difference between the input and restored images, which allows for pixel-wise crack detection. Our comprehensive experimental results demonstrate that UP-CrackNet outperforms other general-purpose unsupervised anomaly detection algorithms, and exhibits satisfactory performance and superior generalizability when compared with state-of-the-art supervised crack segmentation algorithms. Our source code is publicly available at \url{mias.group/UP-CrackNet}.  
\end{abstract}


\IEEEpeerreviewmaketitle
\section{Introduction}
\IEEEPARstart{C}{RACKS} are slender, dark lines or curves that appear on the surface of solid materials, such as roads and bridges \cite{zou2018deepcrack}. Road cracks result from the interplay of water and traffic influences \cite{miller2003distress}, including soil swelling, foundation shifting, traffic overcrowding, premature drying, material expansion and contraction, \etc. Road cracks are not just an inconvenience, they significantly affect the reliability and sustainability of civil infrastructure while posing a significant threat to vehicle conditions and driving safety \cite{liu2019deepcrack}. For instance, in the first two months of 2018, drivers in Chicago submitted 11,706 complaints pertaining to road defects. Furthermore, statistics suggest that substandard road conditions are responsible for nearly one-third of the 33,000 traffic fatalities that occur in the United States annually \cite{fan2018road, ma2022computer}. Therefore, to lower the risk of structural degradation and traffic accidents, frequent road inspection is necessary and essential \cite{li2023roadformer}.
Currently, manual visual inspection is still the dominant method for road crack detection \cite{fan2019road}. The locations of road cracks are recorded routinely by civil engineers or qualified inspectors, the process of which is time-consuming, costly, and hazardous \cite{fan2019pothole, fan2021graph}. For example, New Zealand city councils spent millions of dollars in 2017 detecting and repairing road defects (Christchurch alone spent 525,000 USD) \cite{ma2022computer}. Moreover, the detection results are always qualitative and subjective, as decisions depend entirely on personal opinions and expertise. Owing to these concerns, there is an ever-increasing need to develop automated road condition monitoring methods that can detect road cracks accurately, efficiently, and objectively \cite{fan2021rethinking}.

Before the advent of the deep learning revolution, research in road crack detection was primarily dominated by traditional image processing-based techniques, including edge-based \cite{zhao2015anisotropic, ayenu2008evaluating}, thresholding-based \cite{yamaguchi2008image}, texture analysis-based \cite{hu2010novel}, wavelet-based \cite{zhou2006wavelet}, and minimal path search-based methods \cite{amhaz2016automatic}. While these methods may demonstrate effectiveness in certain simple scenarios, they are often characterized by high computational demands and susceptibility to various environmental factors, with illumination and weather conditions being particularly notable \cite{fan2019crack}. Moreover, the geometric presumptions used in such methods are sometimes impractical, due to the irregular shapes of road cracks \cite{guo10udtiri}. 

Fortunately, with recent advances in deep learning, convolutional neural networks (CNNs) have been extensively employed as feasible methods for automated road crack detection. Rather than setting explicit parameters and using hand-crafted features, CNNs are typically trained to update the implicit parameters of neural layers through back-propagation with a huge amount of human-annotated road data. Such data-driven algorithms are commonly divided into three categories: (1) image classification networks, (2) object detection networks, and (3) semantic segmentation networks. The image classification networks \cite{krizhevsky2017imagenet} are trained to distinguish positive (crack) and negative (non-crack) road images \cite{fan2021deep}. Object detection networks are trained to identify road cracks at the instance level (location and class) \cite{cha2018autonomous, du2021pavement}. Semantic segmentation networks \cite{dung2019autonomous, huyan2020cracku, qu2021crack, chen2020pavement, liu2020automated, zhang2022intelligent, zhu2023lightweight} are trained to achieve pixel-wise crack detection results, and they have emerged as the preferred choice for this task in recent years. 

Nonetheless, the aforementioned pixel-wise road crack detection algorithms predominantly rely on supervised learning. On one hand, training these data-driven algorithms demands a large amount of pixel-level human-annotated labels. The annotation process is exceptionally labor-intensive and time-consuming. Moreover, unique road cracks are not ubiquitous, which adds complexity to the task of gathering a sufficient number of images containing road cracks. On the other hand, supervised learning-based algorithms often demonstrate limited generalizability when applied to different scenarios due to their dependency on fixed, pre-defined patterns learned from specific training data, which may not adequately represent the variability and complexity of real-world situations.

To overcome these limitations, we propose an \uline{\textbf{U}nsupervised \textbf{P}ixel-wise \textbf{C}rack Detection \textbf{Net}work (\textbf{UP-CrackNet})} via adversarial image restoration. In the training phase, multi-scale square masks are first generated and selected randomly to corrupt input undamaged road images. 
These corrupted images are subsequently fed into the proposed model, which learns semantic context from surrounding uncorrupted regions to restore the corrupted regions while adhering to a global consistency constraint. In the testing phase, when provided with a damaged road image, the trained model can restore undamaged regions but may not effectively restore crack regions to their original appearance. Consequently, we can obtain an error map by comparing the difference between the input damaged image and the restored image. This error map can then be used to produce pixel-wise crack detection results. We conduct experiments on three public road crack detection datasets. The results suggest that UP-CrackNet can eliminate the need for human annotations during training while outperforming other unsupervised anomaly detection algorithms. Furthermore, it achieves satisfactory performance and shows superior generalizability when compared to state-of-the-art (SoTA) supervised crack detection approaches. Our main contributions are summarized as follows:
\begin{enumerate}
    \item We propose UP-CrackNet, a novel unsupervised network for pixel-wise road crack detection via adversarial image restoration. It uses only undamaged road images in the training phase without any human-annotated labels. 
    \item We design multi-scale square masks to randomly corrupt input undamaged images, which can prevent the network from degenerating into an identity mapping in the inference phase.
    \item We design comprehensive loss functions, enabling the network to learn semantic context features from uncorrupted undamaged regions to restore the corrupted regions.   
    \item We conduct extensive experiments and compare our method with 11 supervised methods and two unsupervised methods. The results suggest that UP-CrackNet outperforms other unsupervised methods and demonstrates satisfactory performance and superior generalizability compared to supervised methods.
\end{enumerate}

The remainder of this article is organized as follows: Sect. \ref{sec.review} reviews related works. Sect. \ref{sec:methods} provides a detailed description of our proposed network and loss functions. Sect. \ref{sec.exp} presents implementation details, evaluation metrics, experimental results, and visualization analysis. Sect. \ref{sec.discussion} discusses failure cases and Sect. \ref{sec.conclusion} concludes the article.

\begin{figure*}[!t]
	\begin{center}
		\centering
		\includegraphics[width=1\textwidth]{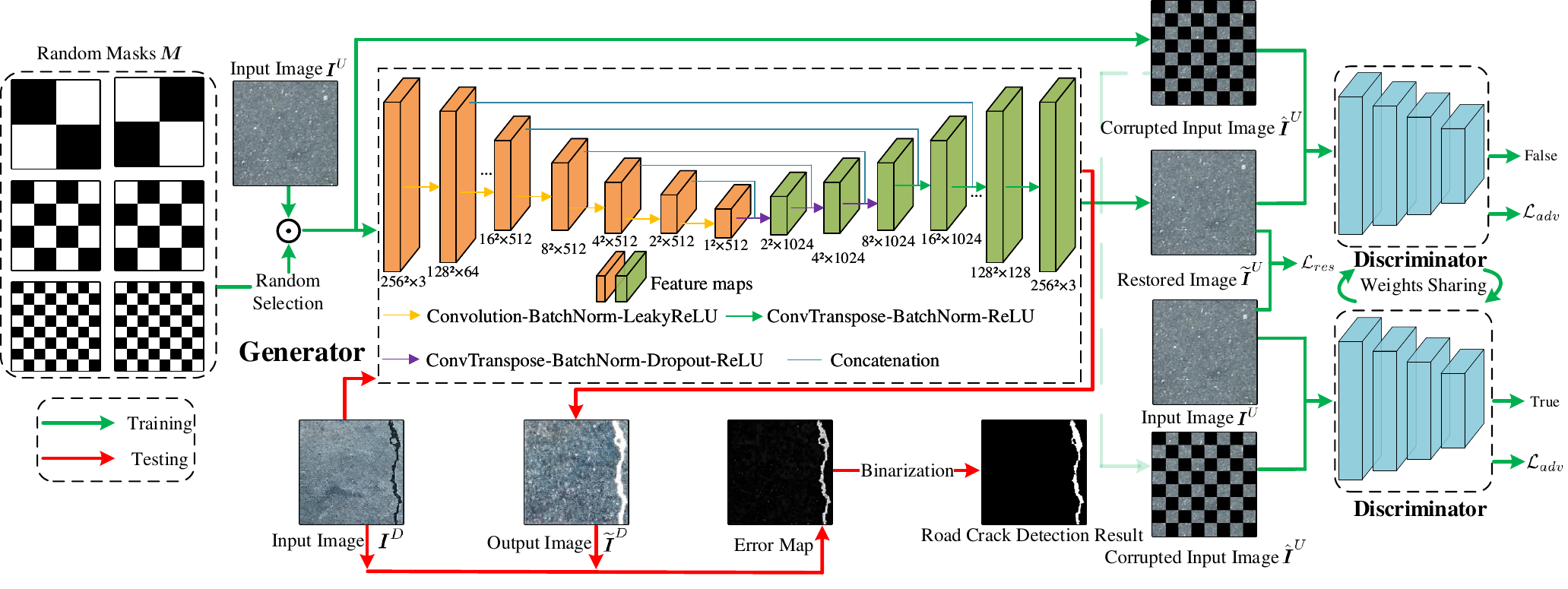}
		\centering
        \captionsetup{font={small}}
		\caption{An illustrative pipeline of our proposed UP-CrackNet.}
		\label{fig1_algorithm}
	\end{center}
\end{figure*}

\section{Literature review}
\label{sec.review}
\subsection{Traditional Road Crack Detection Methods}

Traditional road crack detection methods are generally based on visual features, with edges being a common choice. For instance, bi-dimensional empirical mode decomposition along with the Sobel edge detector was used in \cite{ayenu2008evaluating} to identify road cracks. Thresholding methods are also prevalently employed for this task. Assuming that road cracks consist of thin interconnected textures, in \cite{yamaguchi2008image}, crack textures were extracted by analyzing the connectivity of luminance and shape within the infiltrated regions. Texture analysis-based techniques are another alternative. In \cite{hu2010novel}, local binary pattern operators were utilized to group road patterns and extract distinctive local features for crack detection. Additionally, wavelet-based approaches \cite{zhou2006wavelet, subirats2006automation} decompose road images into different frequency sub-bands to enable the identification of road cracks.
Minimal path search-based methods \cite{amhaz2014new, amhaz2016automatic} are also a popular choice for road crack detection. These methods begin by identifying relatively dark pixels as the endpoints of road cracks and subsequently compute minimal paths between them using path planning techniques to generate road cracks. Nevertheless, the aforementioned traditional road crack detection methods are sensitive to environmental factors and may occasionally prove ineffective, particularly when faced with irregularly shaped road cracks.

\subsection{Supervised Road Crack Detection Methods}

CNNs developed for general computer vision tasks, such as image classification, object detection, and semantic segmentation, have been widely adopted in road crack detection. Image classification networks are employed to categorize road image patches as either negative (indicating healthy road surfaces) or positive (indicating patches containing cracks) \cite{fan2019crack, fan2021deep}, while object detection networks further localize road cracks with bounding boxes \cite{du2021pavement, tsuchiya2019method, hu2021pavement}. Although these networks are unable to produce pixel-wise results, they can be utilized in the preparation of datasets for our proposed unsupervised road crack detection framework.

Semantic segmentation networks, trained through supervised learning, have the capacity to generate pixel-wise road crack detection results. Deepcrack \cite{liu2019deepcrack} incorporates a side-output layer into the VGG-16 \cite{simonyan2014very} model and utilizes conditional random fields and guided filtering to achieve accurate road crack detection results. Another Deepcrack version, proposed in \cite{zou2018deepcrack}, fuses features from various scales of SegNet \cite{badrinarayanan2017segnet} to acquire hierarchical information, leading to improved road crack segmentation performance. RHA-Net \cite{zhu2022rha} integrates residual blocks and hybrid attention modules into an encoder-decoder network for pixel-wise road crack detection. Similarly, DMA-Net \cite{sun2022dma} integrates a multi-scale attention module into the decoder of Deeplabv3+ \cite{chen2018encoder} to dynamically adjust weights across different feature maps for better crack detection results. However, training such methods requires a large amount of human-annotated pixel-level annotations, which is a highly labor-intensive and time-consuming process. Moreover, they often struggle with poor generalizability in unseen datasets. 

\subsection{Unsupervised Anomaly Detection Methods Based on Image Restoration}

Unsupervised anomaly detection approaches based on image restoration have been prevalently used for industrial defect detection. These methods can be categorized into autoencoder (AE)-based \cite{yang2019multiscale, shi2021unsupervised}, variational autoencoder (VAE)-based \cite{dehaene2019iterative, zhou2021semi}, and generative adversarial network (GAN)-based \cite{schlegl2019f}. Among them, GAN-based approaches generate images with the highest quality. However, these methods often generalize to abnormal samples or even degenerate into an identity mapping during the inference phase. To overcome this limitation, researchers attempted to introduce perturbations \cite{sabokrou2018adversarially, ye2020attribute, yan2021learning, zavrtanik2021reconstruction}, which help maintain the dissimilarity between model inputs and outputs, thereby improving the learning of contextual information from normal samples. In \cite{yan2021learning}, a semantic context-based anomaly detection network (SCADN) based on striped masks was proposed. It removes specific regions from the input images and trains a GAN model to restore the corrupted regions. \cite{zavrtanik2021reconstruction} proposed a reconstruction-by-inpainting anomaly detection (RIAD) method, using jumbled small square masks to randomly remove regions of input images and training an AE model with U-Net architecture to restore the corrupted regions. Drawing inspiration from these approaches, we introduce UP-CrackNet, a novel unsupervised pixel-wise road crack detection approach based on adversarial image restoration. UP-CrackNet is trained on undamaged road images during the training phase, without the reliance on any human-annotated labels.

\section{Methodology}
\label{sec:methods}

\subsection{Architecture Overview}

The training and testing processes of our proposed UP-CrackNet are illustrated in Fig. \ref{fig1_algorithm}. During the training phase, we first create corrupted images by performing the Hadamard product operation between the input undamaged road images and randomly generated square masks (where mask values are set to either 0 or 1). Subsequently, we train the proposed model to restore the corrupted regions by minimizing a restoration loss and an adversarial loss. In the testing phase, when provided with damaged road images, the model generates restored images using the learned parameters. Error maps are then obtained by computing the differences between the damaged road images and the restored images. Finally, post-processing techniques are applied to these error maps to enhance the crack detection results. 

\subsection{Undamaged Road Image Random Corruption}
\label{undamage_corrupt}

When designing masks for the random corruption of input undamaged road images, we take into consideration the need for image regions to have an equal chance of being removed. This ensures that all undamaged regions in the training set have an equal probability of being learned by the model.
Specifically, an image is divided into $\frac{H}{k} \times \frac{W}{k}$ patches, where $H$ and $W$ represent the height and width of input undamaged road images, respectively, and $k$ determines the density of patches. We use a boolean logic strategy to design masks, where pixel values are set to 0 or 1 to indicate the regions that should be removed or retained, respectively. The ratio between the removed and retained regions is $1:1$. Given the undamaged road crack training set $\mathcal{I}^U$, the random corruption process can be formulated as $\hat{\boldsymbol{I}}^{U} = \boldsymbol{I}^U \odot \boldsymbol{M}$, where $\boldsymbol{I}^U \in \mathcal{I}^U$ denotes the input undamaged road image, $\boldsymbol{M}$ denotes the selected mask, $\odot$ denotes the Hadamard product, and ${\hat{\boldsymbol{I}}^U}$ denotes the corrupted input undamaged image. 

\subsection{Adversarial Image Restoration}
\label{adversarial_image}
Our proposed UP-CrackNet consists of a generator $G$ and a discriminator $D$. $G$ is trained to restore the corrupted regions by minimizing a restoration loss and an adversarial loss, while $D$ is designed to discriminate between input undamaged road images and the restored images generated by $G$, with the aim of maximizing the adversarial loss.

Generator $G$ consists of an encoder and a decoder. The encoder uses modules of the form \texttt{Convolution-BatchNorm-LeakyReLU}, where the \texttt{BatchNorm} layer performs normalization for each mini-batch to expedite training, and the \texttt{LeakyReLU} layer prevents the vanishing gradient problem by providing small-slope outputs for negative inputs, ensuring that potentially valuable information is retained. On the other hand, the decoder uses two types of modules of the form \texttt{ConvTranspose-BatchNorm-Dropout-ReLU} and \texttt{ConvTranspose-BatchNorm-ReLU}. The \texttt{Dropout} layer randomly deactivates half of the input units, introducing stochasticity to enhance network generalization.  Additionally, we adopt the U-Net architecture as the backbone of $G$ to learn semantic context information. In a mathematical formulation, $G$ can be represented as $\widetilde{\boldsymbol{I}}^U = Dec[Enc(\hat{\boldsymbol{I}}^U)]$, where $Enc$ denotes the encoder, $Dec$ denotes the decoder, and $\widetilde{\boldsymbol{I}}^U$ denotes the restored image.

Discriminator $D$ also uses modules of the form \texttt{Convolution-BatchNorm-LeakyReLU}. Nevertheless, it takes two pairs of images concatenated together as input. $D$ is trained to distinguish between fake image ($\widetilde{\boldsymbol{I}}^U$ generated from $G$) and real image (the input undamaged road image $\boldsymbol{I}^U$) conditioned on the corrupted image $\hat{\boldsymbol{I}}^U$.

\subsection{Road Crack Detection}

In the testing phase, given a damaged road image $\boldsymbol{I}^D$, the detection results $\boldsymbol{S}$ can be obtained using the following expression:
\begin{equation}
    \boldsymbol{S} = O[B((\widetilde{\boldsymbol{I}}^D - \boldsymbol{I}^D) \odot (\widetilde{\boldsymbol{I}}^D - \boldsymbol{I}^D))],
\end{equation}
where $\widetilde{\boldsymbol{I}}^D$ denotes the restored image, $B$ denotes the bilateral filtering \cite{tomasi1998bilateral} operation used to reduce small incorrectly detected regions for improved road crack detection performance, and $O$ denotes the Otsu's thresholding \cite{otsu1979threshold} operation to binarize error maps for pixel-wise crack detection.

\subsection{Loss Functions}
\label{loss_functions}

The total loss function is as follows:
\begin{equation}
    \label{total_loss}
    \mathcal{L}_{total} = \lambda_{res}\mathcal{L}_{res} + \lambda_{adv}\mathcal{L}_{adv},
\end{equation}
where $\lambda_{res}$ and $\lambda_{adv}$ are hyper-parameters used to balance the restoration loss $\mathcal{L}_{rec}$ and the adversarial loss $\mathcal{L}_{adv}$. $G$ is updated by minimizing $\mathcal{L}_{total}$, while $D$ is updated by maximizing $\mathcal{L}_{adv}$.

\subsubsection{Restoration Loss}

We use the mean average error (MAE) loss to measure the difference between $\boldsymbol{I}^U$ and $\widetilde{\boldsymbol{I}}^U$:
\begin{equation}
   \mathcal{L}_{MAE} = \Vert \widetilde{\boldsymbol{I}}^U - \boldsymbol{I}^U \Vert _1.
\end{equation}
However, the MAE loss calculates the pixel intensity differences independently, ignoring the correlation between neighboring pixels. Therefore, we also use a structured similarity index measure (SSIM) loss \cite{wang2004image} and a multi-scale gradient magnitude similarity (MSGMS) loss \cite{zavrtanik2021reconstruction} to measure the structural difference between them:
\begin{equation}
    \mathcal{L}_{SSIM} = 
    \frac{1}{H \times W} \sum\limits_{i=1}\limits^{H} \sum\limits_{j=1}\limits^{W}[1 - SSIM(\boldsymbol{I}^U, \widetilde{\boldsymbol{I}}^U)_{(i,j)}],
\end{equation}
\begin{equation}
    \mathcal{L}_{MSGMS} = \frac{1}{4} \sum\limits_{l=1}\limits^{4}
    \frac{1}{N_{l}} \sum\limits_{i=1}\limits^{H_{l}} \sum\limits_{j=1}\limits^{W_{l}}[1 - GMS(\boldsymbol{I}^U, \widetilde{\boldsymbol{I}}^U)_{(i,j)}],
\end{equation}
where $SSIM$ refers to the SSIM value \cite{wang2004image} between two patches of $\boldsymbol{I}^U$ and $\widetilde{\boldsymbol{I}}^U$ centered at pixel $(i,j)$. The MSGMS loss is calculated over an image pyramid of four different scales, including the original image, and images that are $\frac{1}{2}$, $\frac{1}{4}$, and $\frac{1}{8}$ of the original size. $H_{l}$ and $W_{l}$ represent the height and width of the image at scale $l$, respectively, and $N_{l}$ denotes the number of pixels at scale $l$, respectively. $GMS$ refers to the value of GMS map \cite{xue2013gradient} of $\boldsymbol{I}^U$ and $\widetilde{\boldsymbol{I}}^U$ at pixel $(i,j)$.
Additionally, we employ a style loss \cite{johnson2016perceptual} as follows:
\begin{equation}
    \mathcal{L}_{style} = \mathbb{E}_{i}[\lvert G_{i}^{\phi}(\widetilde{\boldsymbol{I}}^U) - G_{i}^{\phi}(\boldsymbol{I}^U) \lvert]
\end{equation}
to measure the feature difference between $\boldsymbol{I}^U$ and $\widetilde{\boldsymbol{I}}^U$, where $G_{i}^{\phi}$ represents a $C_{i} \times C_{i}$ gram matrix constructed from ${\phi}_{i}$, which denotes the activation map of the $i-$th layer of the pre-trained network.
Therefore, the total restoration loss is formulated as follows:
\begin{equation}
    \begin{aligned}
    \mathcal{L}_{res} = & \lambda_{mae}\mathcal{L}_{MAE} + \lambda_{ssim}\mathcal{L}_{SSIM} + 
    \\
    & \lambda_{gms}\mathcal{L}_{MSGMS} + \lambda_{style}\mathcal{L}_{style},
    \end{aligned}
\end{equation}
where $\lambda_{mae}$, $\lambda_{ssim}$, $\lambda_{gms}$ and $\lambda_{style}$ are hyper-parameters used to balance these losses.

\subsubsection{Adversarial loss}
The adversarial loss is formulated as follows:
\begin{equation}
    \begin{aligned}
        \mathcal{L}_{adv}(G,D)  = & \mathbb{E}_{\hat{\boldsymbol{I}^U}, \boldsymbol{I}^U}[\textup{log}D(\hat{\boldsymbol{I}^U}, \boldsymbol{I}^U)] + \\
    &\mathbb{E}_{\hat{\boldsymbol{I}}^U,z}[\textup{log}(1-D(\hat{\boldsymbol{I}}^U,G(\hat{\boldsymbol{I}}^U,z)],
    \end{aligned}
\end{equation}
where $\widetilde{\boldsymbol{I}}^U = G(\hat{\boldsymbol{I}}^U,z)$ and $z$ denotes random noise introduced by the dropout layers.

\section{Experimental results} 
\label{sec.exp}
\subsection{Datasets}

\begin{table*}[t!] \small
	\renewcommand{\arraystretch}{1}
	\settablefont
	\caption{
 Ablation study results for pixel-wise crack detection performance using five loss functions with the proposed UP-CrackNet on the Crack500 dataset \cite{yang2019feature}. The symbol \Checkmark indicates the selected loss function.
 }
	\centering
	\begin{tabular}{R{0.7cm}R{0.7cm}R{1cm}R{0.7cm}R{0.7cm}R{1.9cm}R{2.0cm}R{2.0cm}R{2.0cm}R{1.8cm}}
		\toprule
		{$\mathcal{L}_{MAE}$}&{$\mathcal{L}_{SSIM}$}&{$\mathcal{L}_{MSGMS}$}&{$\mathcal{L}_{style}$}&{$\mathcal{L}_{adv}$} & Precision~($\%$)$\uparrow$ & Recall~($\%$)$\uparrow$ & Accuracy~($\%$)$\uparrow$ & F1-Score~($\%$)$\uparrow$ & IoU~($\%$)$\uparrow$ \\ \midrule
		{\Checkmark}&{}&{}&{}&{} & 60.026 & 29.387 & 89.704 & 39.457 & 24.577 \\
        {\Checkmark}&{\Checkmark}&{}&{}&{} & 71.705 & 40.326 & 92.488 & 51.621 & 34.790 \\
        {\Checkmark}&{}&{\Checkmark}&{}&{} & 55.188 & 53.561 & 94.821 & 54.363 & 37.327 \\
        {\Checkmark}&{}&{}&{\Checkmark}&{} & 57.074 & 35.620 & 91.835 & 43.864 & 28.093 \\
        {\Checkmark}&{}&{}&{}&{\Checkmark} & 60.876 & 57.502 & 95.299 & 59.141 & 41.986 \\
        {\Checkmark}&{\Checkmark}&{\Checkmark}&{}&{} & 75.341 & 48.014 & 94.063 & 58.651 & 41.494 \\
        {\Checkmark}&{\Checkmark}&{\Checkmark}&{\Checkmark}&{} & \textbf{76.651} & 48.533 & 94.152 & 59.434 & 42.282 \\
        {\Checkmark}&{\Checkmark}&{\Checkmark}&{}&{\Checkmark} & 59.239 & \textbf{61.021} & \textbf{95.607} & 60.116 & 42.976 \\
        {\Checkmark}&{\Checkmark}&{\Checkmark}&{\Checkmark}&{\Checkmark} & 65.377 & 58.609 & 95.484 & \textbf{61.808} & \textbf{44.726} \\
		\bottomrule
		\\
	\end{tabular}
	\label{table7}
\end{table*}

\begin{table}[t!]
	\renewcommand{\arraystretch}{1}
	\settablefont
	\caption{Ablation study results using different modes of masks on the Crack500 dataset \cite{yang2019feature}.}
	\centering
	\begin{tabular}{R{1.7cm}R{1.8cm}R{1.9cm}R{1.5cm}}
		\toprule
		\multicolumn{1}{r} {Mode} & Accuracy~($\%$)$\uparrow$ & F1-Score~($\%$)$\uparrow$ & IoU~($\%$)$\uparrow$ \\ \midrule
        $M_{Jumbled}$ & 94.057 & 48.728 & 32.213\\
        $M_{Striped}$ & 95.379 & 58.932 & 41.775\\
        $M_{Mul-S}$ & \textbf{95.484} & \textbf{61.808} & \textbf{44.726}\\
        \bottomrule
		\\
	\end{tabular}
	\label{table8}
\end{table}

The \textbf{Crack500} \cite{yang2019feature} dataset contains 500 images (resolution: $2,000\times1,500$ pixels) of pavement cracks. These images have been annotated at the pixel level. Each image is cropped into 16 non-overlapped image regions, with only those regions containing more than 1,000 pixels of cracks being retained. This process yields a total of 1,896 training images, 348 validation images, and 1,124 test images. In our experiments, we use the original dataset to train supervised methods. Additionally, we crop 1,896 undamaged road images from the original images to train unsupervised methods. It is important to note that the Crack500 dataset poses significant challenges for practical crack detection, as it includes shadows, occlusions, and varying lighting conditions.

The \textbf{DeepCrack} \cite{liu2019deepcrack} dataset contains $537$ concrete surface images (resolution: $544\times384$ pixels) with multi-scale and multi-scene cracks. These images have also been annotated at the pixel level. The dataset is divided into two subsets, with 300 images used for training and the remaining 237 images used for testing. Similarly, we use the original dataset to train supervised methods, and from the same dataset, we extract 300 undamaged road images to train unsupervised methods.

The \textbf{CFD} \cite{shi2016automatic} dataset contains $118$ concrete surface images (resolution: $480 \times 320$ pixels), manually annotated at the pixel level. These images exhibit diverse illumination conditions, shadows, and stains, making the detection of cracks challenging. We extract 200 image patches (resolution: $256 \times 256$ pixels) to evaluate the generalizability of both supervised and unsupervised methods. 

\subsection{Implementation Details \& Evaluation Metrics}

Our experiments are conducted on a single NVIDIA RTX3090. The models are trained for 200 epochs, with early stopping if there is no performance improvement on the validation set for 20 consecutive epochs. All images are resized to $256 \times 256$ pixels, and data augmentation techniques including scaling, cropping, and flipping are applied. Stochastic gradient descent (SGD) is used to optimize networks, with a momentum value of $0.9$ and weight decay set to $10^{-4}$. 
The initial learning rate is set to $0.01$ and is dynamically adjusted using the poly strategy.
The training settings of UP-CrackNet follow the well-known Pix2Pix \cite{isola2017image}, where an Adam optimizer \cite{kingma2014adam} with $\beta_{1}=0.5$ and $\beta_{2}=0.999$ is used to optimize the networks.
The initial learning rates for the generator $G$ and discriminator $D$ are set to 0.0001 and 0.0004, respectively. During training, the learning rates decay exponentially. The hyper-parameters in the loss functions are set to $\lambda_{mae}=\lambda_{ssim}=\lambda_{gms}=1$, $\lambda_{style}=10$, $\lambda_{res}=100$, and $\lambda_{adv}=1$. In our implementation, $H$ and $W$ are set to 256, and $k$ is set to $\{128, 64, 32\}$. For evaluation, we use precision, recall, accuracy, intersection over union (IoU), and F1-score to quantitatively compare UP-CrackNet with other methods.

\subsection{Ablation Study}
\label{ablation}

\begin{table*}[t!]
	\renewcommand{\arraystretch}{1}
	\settablefont
	\caption{Quantitative experimental results on the Crack500 dataset \cite{yang2019feature}.}
	\centering
    \begin{threeparttable}
	\begin{tabular}{R{2.2cm}R{2.4cm}R{2.0cm}R{2.0cm}R{2.2cm}R{2.0cm}R{2.0cm}}
		\toprule
		\multicolumn{1}{r}{Training Strategy}&{Methods} & Precision~($\%$)$\uparrow$ & Recall~($\%$)$\uparrow$ & Accuracy~($\%$)$\uparrow$ & F1-Score~($\%$)$\uparrow$ & IoU~($\%$)$\uparrow$ \\ \midrule
		\multirow{8}{*}{General Supervised}&{DeepLabv3+ \cite{chen2018encoder}}  & 58.442 & 68.165 & 96.152 & 62.930 & 45.911 \\
		&{ENet \cite{paszke2016enet}} & 57.544 & 65.505 & 95.933 & 61.267 & 44.162 \\
		&{PSPNet \cite{zhao2017pyramid}} & 61.665 & 69.271 & 96.329 & 65.247 & 48.420 \\
        &{UperNet \cite{xiao2018unified}}  & 59.134 & \textbf{72.275} & 96.448 & 65.048 & 48.200 \\
        &{SegResNet \cite{badrinarayanan2017segnet}}  & 58.777 & 64.229 & 95.866 & 61.382 & 44.282\\
		&{UNet \cite{ronneberger2015u}} & 72.465 & 56.611 & 95.357 & 63.565 & 46.589\\
        &{BiSeNetv2 \cite{yu2021bisenet}}  & 69.332 & 64.176 & 96.122 & 66.655 & 49.986 \\
        &{DDRNet \cite{pan2022deep}} & 55.727 & 68.637 & 96.102 & 61.512 & 44.417 \\
        &{Lawin \cite{yan2022lawin}} & 69.413 & 65.919 & 96.284 & 67.621 & 51.081 \\
        \cmidrule(lr){2-7}
        \multirow{2}{*}{Specific Supervised}
        &{Deepcrack19 \cite{liu2019deepcrack}} & \textbf{86.733} & 57.581 & 95.687 & 69.213 & 52.920 \\
        &{Deepcrack18 \cite{zou2018deepcrack}} & 70.356 & 70.919 & \textbf{96.731} & \textbf{70.636} & \textbf{54.603} \\
		\toprule
		\multirow{2}{*}{Unsupervised}&{SCADN \cite{yan2021learning}} & 
		46.723 & 14.064 & 81.064 & 21.620 & 12.120 \\
	    &{RIAD \cite{zavrtanik2021reconstruction}} & 65.420 & 11.665 & 70.374 & 19.799 & 10.987 \\
	    \toprule
		\multirow{1}{*}{Proposed}&{UP-CrackNet} & 65.377 & 58.609 & 95.484 & 61.808 & 44.726 \\
		\bottomrule
        \\
	\end{tabular}
    \end{threeparttable}
	\label{table_c1}
\end{table*}

\begin{table*}[t!]
	\renewcommand{\arraystretch}{1}
	\settablefont
	\caption{Quantitative experimental results on the Deepcrack dataset \cite{liu2019deepcrack}.}
	\centering
    \begin{threeparttable}
	\begin{tabular}{R{2.2cm}R{2.4cm}R{2.0cm}R{2.0cm}R{2.2cm}R{2.0cm}R{2.0cm}}
		\toprule
		\multicolumn{1}{r}{Training Strategy}&{Methods} & Precision~($\%$)$\uparrow$ & Recall~($\%$)$\uparrow$ & Accuracy~($\%$)$\uparrow$ & F1-Score~($\%$)$\uparrow$ & IoU~($\%$)$\uparrow$ \\ \midrule
		\multirow{9}{*}{General Supervised}&{DeepLabv3+ \cite{chen2018encoder}}  & 68.078 & 79.690 & 97.842 & 73.428 & 58.013 \\
		&{ENet \cite{paszke2016enet}} & 66.618 & 82.529 & 97.921 & 73.725 & 58.385 \\
		&{PSPNet \cite{zhao2017pyramid}} & 44.872 & 85.280 & 97.247 & 58.803 & 41.646 \\
        &{UperNet \cite{xiao2018unified}}  & 67.001 & 82.344 & 97.926 & 73.884 & 58.585 \\
        &{SegResNet \cite{badrinarayanan2017segnet}}  & 37.615 & 82.334 & 96.915 & 51.639 & 34.806\\
		&{UNet \cite{ronneberger2015u}} & 46.530 & 88.690 & 97.399 & 61.038 & 43.924\\
        &{BiSeNetv2 \cite{yu2021bisenet}}  & 64.428 & 79.970 & 97.736 & 71.363 & 55.476 \\
        &{DDRNet \cite{pan2022deep}} & 45.270 & 62.123 & 96.395 & 52.374 & 35.478 \\
        &{Lawin \cite{yan2022lawin}} & 62.249 & 84.282 & 97.839 & 71.609 & 55.774 \\
        \cmidrule(lr){2-7}
        \multirow{2}{*}{Specific Supervised}
        &{Deepcrack19 \cite{liu2019deepcrack}} & 88.785 & 68.923 & 97.756 & 77.603 & 63.403 \\
        &{Deepcrack18 \cite{zou2018deepcrack}} & 71.720 & 88.403 & \textbf{98.350} & \textbf{79.192} & \textbf{65.552} \\
		\toprule
		\multirow{2}{*}{Unsupervised}&{SCADN \cite{yan2021learning}} & 
		66.879 & 34.128 & 92.897 & 45.194 & 29.194 \\
	    &{RIAD \cite{zavrtanik2021reconstruction}} & \textbf{90.079} & 20.742 & 84.494 & 33.720 & 20.279 \\
	    \toprule
		\multirow{1}{*}{Proposed}&{UP-CrackNet} & 63.412 & \textbf{88.852} & 98.049 & 74.006 & 58.738 \\
		\bottomrule
	\end{tabular}     
    \end{threeparttable} 
	\label{tablec_2}
\end{table*}

\begin{figure*}[!t]
	\begin{center}
		\centering
		\includegraphics[width=0.99\textwidth]{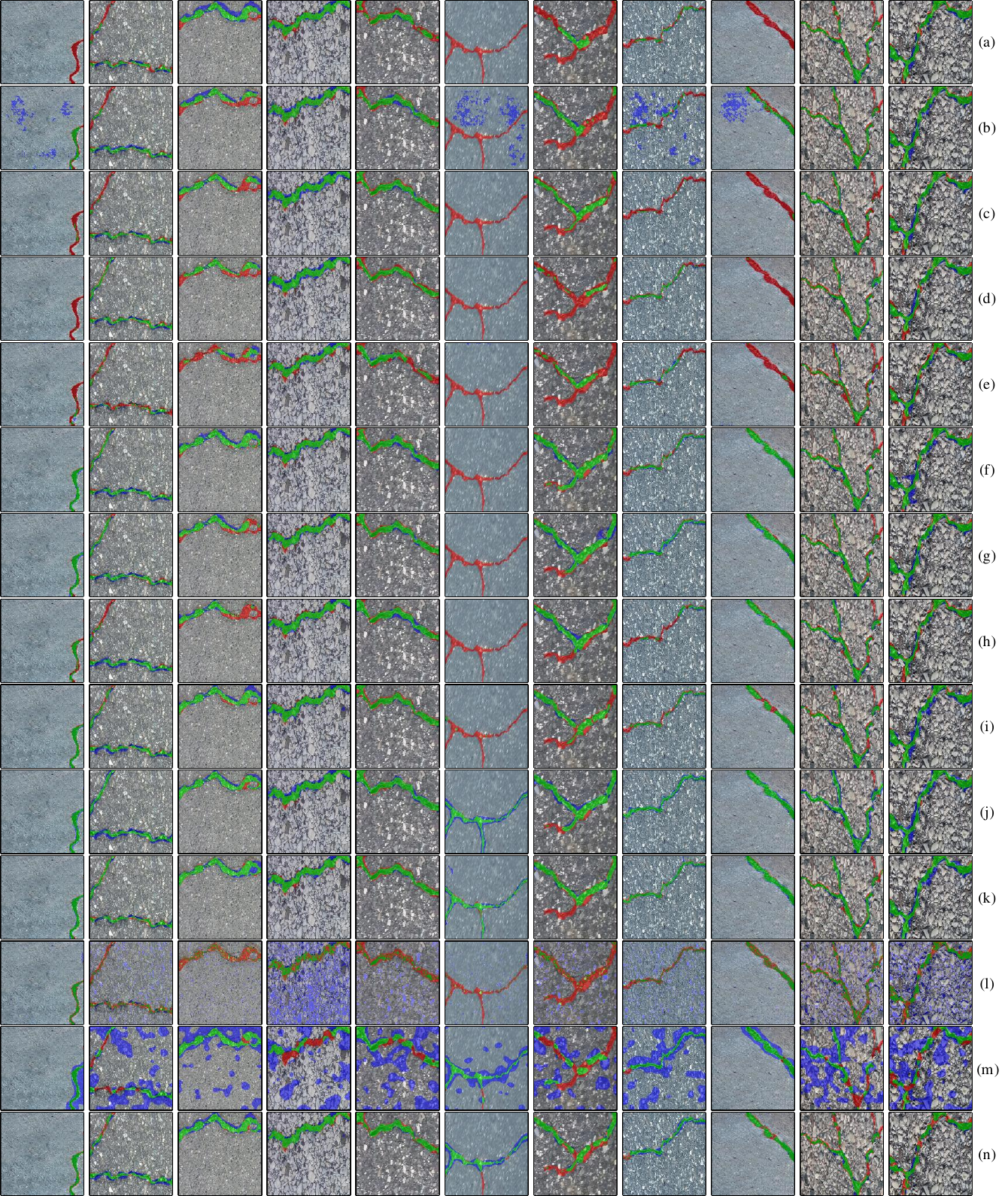}
		\centering
        \settablefont
		\caption{Examples of experimental results on the Crack500 dataset: (a) DeepLabv3+ \cite{chen2018encoder}; (b) ENet \cite{paszke2016enet}; (c) PSPNet \cite{zhao2017pyramid}; (d) UperNet \cite{xiao2018unified}; (e) SegResNet \cite{badrinarayanan2017segnet}; (f) UNet \cite{ronneberger2015u}; (g) BiSeNetv2 \cite{yu2021bisenet}; (h) DDRNet \cite{pan2022deep}; (i) Lawin \cite{yan2022lawin}; (j) Deepcrack19 \cite{liu2019deepcrack}; (k) Deepcrack18 \cite{zou2018deepcrack}; (l) SCADN \cite{yan2021learning}; (m) RIAD \cite{zavrtanik2021reconstruction}; (n) UP-CrackNet. The true-positive, false-positive, and false-negative pixels are shown in green, blue, and red, respectively.}
		\label{fig4}
	\end{center}
\end{figure*}

To analyze the effectiveness of the five employed losses, we conduct an ablation study on the Crack500 dataset. The quantitative results presented in Table \ref{table7} suggest that each loss contributes to improved road crack detection results, and our method, incorporating all these losses, achieves the best performance. Notably, the adversarial loss $\mathcal{L}_{adv}$ provides the most significant improvement among these losses.

To validate the effectiveness of the designed multi-scale square masks, we conduct another ablation study on the same dataset using different modes of masks. The comparison results are given in Table \ref{table8}, where $M_{Mul-S}$ denotes our designed multi-scale square masks, $M_{Striped}$ denotes striped masks used in SCADN, and $M_{Jumbled}$ denotes jumbled small square masks used in RIAD. These results indicate that our method achieves improvements of $2.876\%$ and $13.080\%$ in F1-Score, as well as improvements of $2.951\%$ and $12.513\%$ in IoU compared to using the other two types of masks.

\subsection{Comparison with Other SoTA Methods}
\label{sec:experiments}

\begin{figure*}[!t]
	\begin{center}
	\centering
	\includegraphics[width=\textwidth]{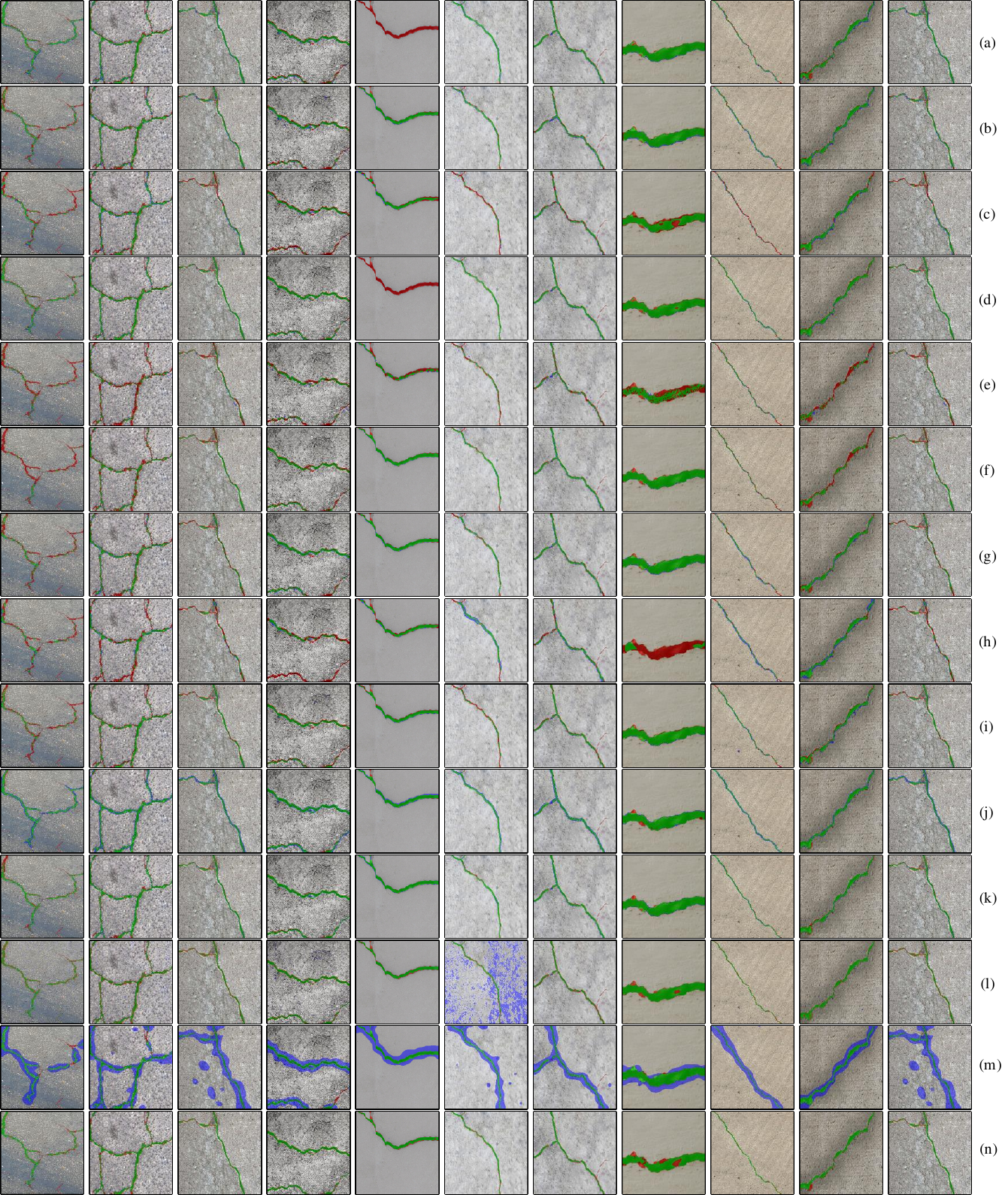}
	\centering
	\caption{Examples of experimental results on the Deepcrack dataset \cite{liu2019deepcrack}:(a) DeepLabv3+ \cite{chen2018encoder}; (b) ENet \cite{paszke2016enet}; (c) PSPNet \cite{zhao2017pyramid}; (d) UperNet \cite{xiao2018unified}; (e) SegResNet \cite{badrinarayanan2017segnet}; (f) UNet \cite{ronneberger2015u}; (g) BiSeNetv2 \cite{yu2021bisenet}; (h) DDRNet \cite{pan2022deep}; (i) Lawin \cite{yan2022lawin}; (j) Deepcrack19 \cite{liu2019deepcrack}; (k) Deepcrack18 \cite{zou2018deepcrack}; (l) SCADN \cite{yan2021learning}; (m) RIAD \cite{zavrtanik2021reconstruction}; (n) UP-CrackNet. The true-positive, false-positive, and false-negative pixels are shown in green, blue, and red, respectively.}
		\label{fig5}
	\end{center}
\end{figure*}

\begin{table*}[t!]
	\renewcommand{\arraystretch}{1}
	\settablefont
	\caption{Quantitative experimental results on the CFD dataset \cite{shi2016automatic} (all trained on the Crack500 dataset \cite{yang2019feature}).}
	\centering
    \begin{threeparttable}
	\begin{tabular}{R{2.2cm}R{2.4cm}R{2.0cm}R{2.0cm}R{2.2cm}R{2.0cm}R{2.0cm}}
		\toprule
		\multicolumn{1}{r}{Training Strategy}&{Methods} & Precision~($\%$)$\uparrow$ & Recall~($\%$)$\uparrow$ & Accuracy~($\%$)$\uparrow$ & F1-Score~($\%$)$\uparrow$ & IoU~($\%$)$\uparrow$ \\ \midrule
		\multirow{9}{*}{General Supervised}&{DeepLabv3+ \cite{chen2018encoder}}  & 6.736  & 54.663 & 98.277 & 11.994 & 6.380 \\
		&{ENet \cite{paszke2016enet}} & 0.123 & 0.206 & 98.261 & 0.156 & 0.078 \\
		&{PSPNet \cite{zhao2017pyramid}} & 11.364 & 52.256 & 98.274 & 18.668 & 10.295 \\
		&{UperNet \cite{xiao2018unified}}  & 7.174 &  \textbf{55.643} & \textbf{98.282} & 12.709 & 6.786\\
        &{SegResNet \cite{badrinarayanan2017segnet}}  & 0.088 & 0.739 & 98.052 & 0.157 & 0.079\\
		&{UNet \cite{ronneberger2015u}} & 15.674 & 48.025 & 98.235 & 23.634 & 13.401\\
        &{BiSeNetv2 \cite{yu2021bisenet}}  & 18.85 & 21.058 & 97.353 & 19.892 & 11.045 \\
        &{DDRNet \cite{pan2022deep}} & 0.084 & 31.993 & 98.255 & 0.167 & 0.083 \\
        &{Lawin \cite{yan2022lawin}} & 38.508 & 37.250 & 97.797 & 37.869 & 23.357 \\
        \cmidrule(lr){2-7}
        \multirow{2}{*}{Specific Supervised}
        &{Deepcrack19 \cite{liu2019deepcrack}} & \textbf{95.215} & 29.446 & 95.940 & 44.981 & 29.017 \\
        &{Deepcrack18 \cite{zou2018deepcrack}} & 77.500 & 43.759 & 97.872 & \textbf{55.935} & \textbf{38.826} \\
		\toprule
		\multirow{2}{*}{Unsupervised}&{SCADN \cite{yan2021learning}} & 
 		43.424 & 15.343 & 94.837 & 22.674 & 12.787 \\
	    &{RIAD \cite{zavrtanik2021reconstruction}} & 93.619 & 11.553 & 87.395 & 20.568 & 11.463 \\
	    \toprule
		\multirow{1}{*}{Proposed}&{UP-CrackNet} & 63.161 & 41.253 & 97.790 & 49.909 & 33.252 \\
		\bottomrule
		\\
	\end{tabular}
    \end{threeparttable}
	\label{table3}
\end{table*}

\begin{table*}[t!]
	\renewcommand{\arraystretch}{1}
	\settablefont
	\caption{Quantitative experimental results on the Deepcrack dataset \cite{liu2019deepcrack} (all trained on the Crack500 dataset \cite{yang2019feature}).}
	\centering
    \begin{threeparttable}
	\begin{tabular}{R{2.2cm}R{2.4cm}R{2.0cm}R{2.0cm}R{2.2cm}R{2.0cm}R{2.0cm}}
		\toprule
		\multicolumn{1}{r}{Training Strategy}&{Methods} & Precision~($\%$)$\uparrow$ & Recall~($\%$)$\uparrow$ & Accuracy~($\%$)$\uparrow$ & F1-Score~($\%$)$\uparrow$ & IoU~($\%$)$\uparrow$ \\ \midrule
		\multirow{9}{*}{General Supervised}&{DeepLabv3+ \cite{chen2018encoder}}  & 41.074 & 32.452 & 93.716 & 36.258 & 22.143 \\
		&{ENet \cite{paszke2016enet}} & 63.108 & 37.842 & 93.884 & 47.313 & 30.987 \\
		&{PSPNet \cite{zhao2017pyramid}} & 31.923 & 73.212 & 96.529 & 44.460 & 28.584 \\
        &{UperNet \cite{xiao2018unified}}  & 29.161 & 66.696 & 96.284 & 40.579 & 25.454\\
		&{SegResNet \cite{badrinarayanan2017segnet}}  & 50.286 & 36.141 & 93.971 & 42.056 & 26.627\\
		&{UNet \cite{ronneberger2015u}} & 79.396 & 22.930 & 87.492 & 35.583 & 21.642\\
        &{BiSeNetv2 \cite{yu2021bisenet}}  & 76.207 & 20.107 & 85.795 & 31.819 & 18.920 \\
        &{DDRNet \cite{pan2022deep}} & 38.233 & 64.599 & 96.402 & 48.036 & 31.610 \\
        &{Lawin \cite{yan2022lawin}} & 54.994 & 63.056 & 96.641 & 58.749 & 41.592 \\
        \cmidrule(lr){2-7}
        \multirow{2}{*}{Specific Supervised}
        &{Deepcrack19 \cite{liu2019deepcrack}} & \textbf{95.828} & 48.461 & 95.384 & 64.370 & 47.460 \\
        &{Deepcrack18 \cite{zou2018deepcrack}} & 90.349 & 61.837 & 97.154 & 73.422 & 58.006 \\
		\toprule
		\multirow{2}{*}{Unsupervised}&{SCADN \cite{yan2021learning}} & 59.006 & 16.970 & 85.660 & 26.360 & 15.181 \\
	    &{RIAD \cite{zavrtanik2021reconstruction}} & 86.545 & 16.620 & 80.529 & 27.884 & 16.201 \\
	    \toprule
		\multirow{1}{*}{Proposed}&{UP-CrackNet} & 72.464 & \textbf{79.531} & \textbf{97.990} & \textbf{75.833} & \textbf{61.073} \\
		\bottomrule
		\\
	\end{tabular}     
    \end{threeparttable}
	\label{table4}
\end{table*}

We compare our proposed UP-CrackNet with 11 general supervised semantic segmentation methods, two supervised crack detection-specific methods, and two general image restoration-based unsupervised anomaly detection methods on the Crack500 and Deepcrack datasets. The quantitative and qualitative comparison results are shown in Table \ref{table_c1}, Table \ref{tablec_2}, Fig. \ref{fig4} and Fig. \ref{fig5}, respectively. It can be observed that UP-CrackNet achieves satisfactory detection performance compared with supervised methods and performs much better than other general image restoration-based unsupervised methods.  

Specifically, on the Crack500 dataset, our proposed UP-CrackNet has $1.185\%-6.355\%$ and $8.194\%-9.877\%$ reduction in IoU than the general and crack detection-specific supervised methods, respectively. On the Deepcrack dataset \cite{liu2019deepcrack}, UP-CrackNet achieves better performance than all general supervised methods and has $4.665\%-6.814\%$ reduction in IoU than specific supervised methods. Considering supervised methods require a large amount of human-annotated pixel-level labels, our proposed UP-CrackNet with satisfactory detection performance has strong application prospects. Compared with two unsupervised methods, our proposed UP-CrackNet achieves an increase of $32.606\%-33.739\%$ and $29.544\%-38.459\%$ IoU improvement on the Crack500 dataset \cite{yang2019feature} and Deepcrack dataset \cite{liu2019deepcrack}, respectively. These results demonstrate that our designed model and losses are more effective for unsupervised pixel-wise crack detection.

\begin{table*}[t!]
	\renewcommand{\arraystretch}{1}
	\settablefont
	\caption{Quantitative experimental results on the CFD dataset \cite{shi2016automatic} (all trained on the Deepcrack dataset \cite{liu2019deepcrack}).}
	\centering
    \begin{threeparttable}
	\begin{tabular}{R{2.2cm}R{2.4cm}R{2.0cm}R{2.0cm}R{2.2cm}R{2.0cm}R{2.0cm}}
		\toprule
		\multicolumn{1}{r}{Training Strategy}&{Methods} & Precision~($\%$)$\uparrow$ & Recall~($\%$)$\uparrow$ & Accuracy~($\%$)$\uparrow$ & F1-Score~($\%$)$\uparrow$ & IoU~($\%$)$\uparrow$ \\ \midrule
		\multirow{9}{*}{General Supervised}&{DeepLabv3+ \cite{chen2018encoder}}  & 37.292 & 48.669 & 98.221 & 42.228 & 26.765 \\
		&{ENet \cite{paszke2016enet}} & 17.826 & 50.529 & 98.263 & 26.354 & 15.177 \\
		&{PSPNet \cite{zhao2017pyramid}} & 22.725 & 42.553 & 98.118 & 29.627 & 17.390 \\
		&{UperNet \cite{xiao2018unified}}  & 31.493 & 51.817 & 98.295 & 39.176 & 24.360\\
        &{SegResNet \cite{badrinarayanan2017segnet}}  & 16.102 & 48.689 & 98.242 & 24.200 & 13.766\\
		&{UNet \cite{ronneberger2015u}} & 25.609 & 53.281 & 98.312 & 34.592 & 20.913\\
        &{BiSeNetv2 \cite{yu2021bisenet}}  & 15.706 & 49.193 & 98.248 & 23.810 & 13.514 \\
        &{DDRNet \cite{pan2022deep}} & 9.153 & 46.625 & 98.234 & 15.301 & 8.285 \\
        &{Lawin \cite{yan2022lawin}} & 14.903 & \textbf{55.158} & 98.305 & 23.466 & 13.292 \\
        \cmidrule(lr){2-7}
        \multirow{2}{*}{Specific Supervised}
        &{Deepcrack19 \cite{liu2019deepcrack}} & 77.205 & 38.407 & 97.444 & \textbf{51.296} & \textbf{34.495} \\
        &{Deepcrack18 \cite{zou2018deepcrack}} & 35.837 & 54.237 & \textbf{98.354} & 43.158 & 27.517 \\
		\toprule
		\multirow{2}{*}{Unsupervised}&{SCADN \cite{yan2021learning}} & 
 		64.689 & 8.626 & 87.439 & 15.222 & 8.238 \\
	    &{RIAD \cite{zavrtanik2021reconstruction}} & \textbf{92.259} & 10.824 & 86.614 & 19.374 & 10.726 \\
	    \toprule
		\multirow{1}{*}{Proposed}&{UP-CrackNet} & 52.388 & 42.267 & 97.923 & 46.787 & 30.537 \\
		\bottomrule
		\\
	\end{tabular}
    \end{threeparttable}
	\label{table5}
\end{table*}

\begin{table*}[t!]
	\renewcommand{\arraystretch}{1}
	\settablefont
	\caption{Quantitative experimental results on the Crack500 dataset \cite{yang2019feature} (all trained on the Deepcrack dataset \cite{liu2019deepcrack}).}
	\centering
    \begin{threeparttable}
	\begin{tabular}{R{2.2cm}R{2.4cm}R{2.0cm}R{2.0cm}R{2.2cm}R{2.0cm}R{2.0cm}}
		\toprule
		\multicolumn{1}{r}{Training Strategy}&{Methods} & Precision~($\%$)$\uparrow$ & Recall~($\%$)$\uparrow$ & Accuracy~($\%$)$\uparrow$ & F1-Score~($\%$)$\uparrow$ & IoU~($\%$)$\uparrow$ \\ \midrule
		\multirow{9}{*}{General Supervised}&{DeepLabv3+ \cite{chen2018encoder}}  & 33.991 & 44.695 & 93.959 & 38.615 & 23.927 \\
		&{ENet \cite{paszke2016enet}} & 15.590 & 54.308 & 94.548 & 24.226 & 13.782 \\
		&{PSPNet \cite{zhao2017pyramid}} & 8.285 & \textbf{85.049} & 94.792 & 15.099 & 8.166 \\
		&{UperNet \cite{xiao2018unified}}  & 46.681 & 42.171 & 93.441 & 44.311 & 28.462\\
        &{SegResNet \cite{badrinarayanan2017segnet}}  & 8.720 & 64.049 & 94.624 & 15.349 & 8.313\\
		&{UNet \cite{ronneberger2015u}} & 5.656 & 47.105 & 94.371 & 10.099 & 5.318\\
        &{BiSeNetv2 \cite{yu2021bisenet}}  & 20.450 & 40.461 & 93.871 & 27.169 & 15.720 \\
        &{DDRNet \cite{pan2022deep}} & 40.761 & 16.881 & 85.470 & 23.875 & 13.555 \\
        &{Lawin \cite{yan2022lawin}} & 20.272 & 76.551 & 95.196 & 32.055 & 19.087 \\
        \cmidrule(lr){2-7}
        \multirow{2}{*}{Specific Supervised}
        &{Deepcrack19 \cite{liu2019deepcrack}} & 61.160 & 51.056 & 94.552 & 55.653 & 38.555 \\
        &{Deepcrack18 \cite{zou2018deepcrack}} & 21.196 & 77.557 & \textbf{95.252} & 33.294 & 19.971 \\
		\toprule
		\multirow{2}{*}{Unsupervised}&{SCADN \cite{yan2021learning}} & 
 		49.131 & 14.896 & 81.466 & 22.861 & 12.906 \\
	    &{RIAD \cite{zavrtanik2021reconstruction}} & \textbf{67.502} & 12.413 & 71.558 & 20.969 & 11.713 \\
	    \toprule
		\multirow{1}{*}{Proposed}&{UP-CrackNet} & 62.796 & 52.200 & 94.706 & \textbf{57.010} & \textbf{39.870}\\
		\bottomrule
		\\
	\end{tabular}           
    \end{threeparttable}
	\label{table6}
\end{table*}

\subsection{Generalizability Evaluation}
To further evaluate the generalizability of the compared methods, we assess the performance of networks trained on the Crack500 dataset and the Deepcrack dataset on other datasets. The quantitative comparison results are shown in Tables \ref{table3}, \ref{table4}, \ref{table5}, and \ref{table6}, respectively. These results indicate that our proposed UP-CrackNet has superior generalizability than other supervised and unsupervised methods. 

When UP-CrackNet is trained on the Crack500 dataset and tested on the Deepcrack dataset, it demonstrates IoU improvements ranging from 3.067\% to 45.892\%. Conversely, when the training and test sets are switched, UP-CrackNet still demonstrates substantial IoU improvements, ranging from 1.315\% to 34.552\%. We attribute these improvements in generalizability to the efficacy of unsupervised image restoration, which allows the model to restore the corrupted regions by analyzing the context provided by surrounding patches. This capacity to capture meaningful semantic information from neighboring contexts significantly enhances UP-CrackNet's performance across different datasets and test scenarios. It is worth noting that although UP-CrackNet performs slightly worse than Deepcrack19 when evaluated on the CFD dataset, we believe this discrepancy is primarily attributed to the dataset itself. The CFD dataset contains small and thin road cracks, making it challenging for image restoration-based algorithms to preserve the clear boundaries of these cracks.

\section{Discussion}
\label{sec.discussion}

As discussed above, it is imperative to divide a given road crack dataset into two sets of road image patches: one containing road cracks and the other free of any cracks. This separation is a fundamental step in training UP-CrackNet, allowing the model to effectively restore undamaged road regions. Fortunately, road crack detection is a relatively easy image classification task, as has been quantitatively demonstrated in \cite{fan2021deep}. Therefore, after extracting image patches from the original images within road crack datasets, pre-trained crack classification networks can be employed to accurately identify undamaged road image patches, which are subsequently fed into UP-CrackNet. 

\begin{figure}[t!]
	\begin{center}
	\includegraphics[width=0.5\textwidth]{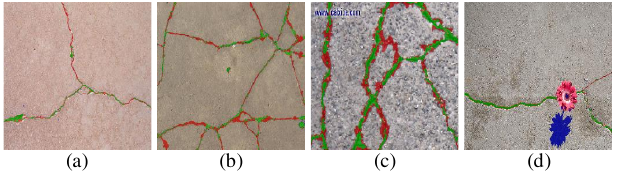}
        \captionsetup{font={small}}
	\caption{Failure cases of UP-CrackNet on the Deepcrack dataset \cite{liu2019deepcrack}. The true-positive, false-positive, and false-negative pixels are shown in green, blue, and red, respectively.}
		\label{fig9}
	\end{center}
\end{figure}

We also present some instances where UP-CrackNet encounters challenges on the Deepcrack dataset \cite{liu2019deepcrack} in Fig. \ref{fig9}. (a) and (b) illustrate situations where UP-CrackNet sometimes does not perform as expected when detecting thin cracks. This can occur because our method may occasionally categorize these small and thin cracks as undamaged regions, resulting in their restoration to their original appearance and, consequently, missed detection. (c) and (d) illustrate cases where UP-CrackNet erroneously identifies image watermark digits and shadows of flowers as damaged areas. This happens because these patterns either do not appear or occur very rarely in the training set. Consequently, UP-CrackNet fails to restore them to their original appearance, leading to false detections. To address these challenges, future work can focus on developing more advanced training mechanisms and network architectures to enhance the detection of thin cracks. Additionally, improving the robustness and intelligence of UP-CrackNet to distinguish between road cracks and other anomalies is an area that warrants further research and development. 

\section{Conclusion}
\label{sec.conclusion}

This article introduced UP-CrackNet, a novel network architecture and training paradigm designed to overcome the limitations of previous supervised pixel-wise road crack detection algorithms. The training of UP-CrackNet was exclusively performed using undamaged road image patches, where an adversarial image restoration technique was applied to learn corrupted regions in an unsupervised manner. The testing process involves a series of conventional image processing algorithms, including bilateral filtering and Otsu's thresholding. Extensive experiments conducted on three datasets demonstrate the effectiveness of our UP-CrackNet in detecting road cracks and its superior generalizability across different datasets and scenarios. In the future, we intend to investigate alternative training strategies and network architectures to further improve UP-CrackNet's performance in detecting thin cracks. Additionally, conducting real-world experiments involving automatic road inspection robots or vehicles is also part of our future research endeavors.

\bibliographystyle{IEEEtran}

\bibliography{main}
\IEEEauthorblockN{\textbf{Nachuan Ma}} 
\IEEEauthorblockA{is currently pursuing his Ph.D. degree at Tongji University with a research focus on visual perception techniques for autonomous driving.}\\
\newline
\IEEEauthorblockN{\textbf{Rui Fan}}
\IEEEauthorblockA{(Senior Member, IEEE) is currently a Full Professor with the College of Electronics \& Information Engineering and the Shanghai Research Institute for Intelligent Autonomous Systems at Tongji University. His research interests include computer vision, deep learning, and robotics.}\\
\newline
\IEEEauthorblockN{\textbf{Lihua Xie}}
\IEEEauthorblockA{(Fellow, IEEE) has been a Professor and Director at the Center for Advanced Robotics Technology Innovation in the School of Electrical and Electronic Engineering at Nanyang Technological University, Singapore. Dr. Xie’s research interests include robust control and estimation, networked control systems, multi-agent networks, and unmanned systems.}

\end{document}